\documentclass[runningheads]{llncs}
\usepackage[T1]{fontenc}
\usepackage{subfig,graphicx}
\usepackage{amsmath,amsfonts}
\usepackage{tabularx}
\usepackage{xcolor}

\begin{document}
\title{A Learning Algorithm for Threshold Boolean Networks with Prescribed Fixed Points}

\titlerunning{Learning TBNs with Prescribed Fixed Points}
%
%
\author{Gonzalo A. Ruz\inst{1,2,3,4}\orcidID{0000-0001-7740-9865}}
\authorrunning{Gonzalo A. Ruz}
%

\institute{Facultad de Ingeniería y Ciencias, Universidad Adolfo Ibáñez, Santiago, Chile \and
Millennium Nucleus for Social Data Science (SODAS), Santiago, Chile \and
Millennium Nucleus in Data Science for Plant Resilience (PhytoLearning), Santiago, Chile \and
Center of Applied Ecology and Sustainability (CAPES), Santiago, Chile
\email{gonzalo.ruz@uai.cl}}
\maketitle              
\begin{abstract}
We present a learning algorithm for inferring threshold Boolean networks (TBNs) with a prescribed set of fixed points. The proposed method employs a custom differentiable loss function that jointly enforces fixed point preservation, penalizes spurious attractors, encourages binary outputs, and promotes sparsity through L1 regularization. Applied to the FOS-GRN model of \emph{Arabidopsis thaliana}, the approach achieved perfect reconstruction (i.e., all 10 desired fixed points and no spurious ones) in 5 out of 30 independent runs, recovering on average 8.53~$\pm$~0.90 correct fixed points with no spurious attractors. In contrast, standard methods such as the Perceptron and Logistic Regression recovered up to 10 fixed points but introduced between 8 and 31 spurious ones. An additional analysis varying the sparsity coefficient ($\lambda_{\ell_1}$) confirmed that the method’s performance and the structural properties of the inferred networks remain robust within a practical range (up to 0.01) of regularization strengths. Overall, the results demonstrate the effectiveness and stability of the proposed algorithm in capturing meaningful network dynamics under prescribed dynamical constraints.

\end{abstract}
\section{Introduction}
\label{sec:SCIENTIFIC-BACKGROUND}
Boolean networks are a common modeling tool to represent gene regulatory networks \cite{Kauffman69,SANCHEZCORRALES2010971,Faure06,Tania2020}. They have also been used to model robots \cite{10.1007/978-3-642-20525-5_5}, consensus mechanisms \cite{9892553,Salva}, cryptography \cite{Zanin_2011}, and the dynamics of industrial networks \cite{10.1007/978-3-540-85081-6_44}. A Boolean network is a directed graph where each node represents a gene (or, more generally, the entity being modeled) that can have two possible values: 1 (active) or 0 (inactive). The edges represent regulatory relations, and the network nodes update their values based on local Boolean functions, one per node, typically using logic gates such as AND, OR, and NOT, applied to the values of the neighboring nodes. For a Boolean network with $n$ nodes, there are $2^{n}$ states or configurations. The network eventually converges to steady states or attractors by repeatedly applying the update rule from any initial configuration. There are two types of attractors; a fixed point is a state where the network remains ``stuck'' in that state forever. A limit cycle is a set of states the network loops around forever. A Boolean network can have several fixed points and limit cycles, but in general, from a modeling point of view, only the fixed points are of interest, as they often correspond to stable phenotypes or cell types; for example, in \cite{Mendoza98}, the Mendoza \& Alvarez-Buylla network that captured the dynamics of the floral development in \textit{Arabidopsis thaliana}. The model consists of twelve interacting chemical species, designated by EMF1, TFL1, LFY, AP1, CAL, LUG, UFO, BFU, AG, AP3, PI, and SUP. The BFU species is a dimer of the AP3 and PI proteins, and all the rest are proteins as well. By starting in any of the $2^{12} = 4096$ possible configurations, and using the parallel updating scheme, the network converges to one of the possible thirteen attractors. Six are the following fixed points: 1) 000100000000, 2) 000100010110, 3) 000000001000, 4) 000000011110, 5) 110000000000, 6) 110000010110. Each one has associated a cell type: 1) sepal, 2) petal, 3) carpel, 4) stamen, 5) inflorescence, 6) mutant (unobserved cell). The remaining seven are limit cycles of length two, which have no biological meaning. Similarly, in \cite{Deal2023}, a Boolean model of the tryptophan operon in \textit{Escherichia coli} is presented. This model comprises eight Boolean variables (nodes) along with two parameters controlling the operon's activation state. Under synchronous updating and various parameter assignments representing different tryptophan concentration levels, the network demonstrates two distinct fixed points: one corresponding to an OFF state (high tryptophan levels) and the other to an ON state (medium to low
tryptophan levels). However, the model also exhibits spurious limit cycles lacking biological significance.

When inferring Boolean networks from data (binarized gene expression) or when enforcing particular characteristics, for example, a set of attractors, a popular choice is the threshold Boolean network variant, where from a reconstruction point of view, the network's parameters that the learning algorithm needs to infer from the data are the weight matrix and threshold vector \cite{10264894,RUZ2023104902,RUZ2025105572}.

In this context, a recent work \cite{10944722} presented a training approach to infer threshold Boolean networks using the classical Perceptron learning algorithm. The method was applied to infer a threshold Boolean network model of the \textit{Arabidopsis thaliana} flower organ specification gene regulatory network (FOS-GRN), consisting of thirteen genes. The results showed that three genes exhibited non-linear
interactions, which a threshold Boolean network approach could not capture. Nevertheless, the network's asymptotic behavior correctly identified biologically meaningful fixed points with the largest basins
of attraction. When the goal of fully replicating the FOS-GRN
state transition table was relaxed, and instead, the focus shifted
to inferring networks with the ten important fixed points, the proposed method was capable of succeeding in this objective but it constantly introduced a significant number of undesirable spurious fixed points.

This paper introduces a learning algorithm for inferring threshold Boolean networks with prescribed fixed points. For this, a specially designed loss function is proposed, and the training algorithm using the loss function is derived. We compare the proposed method with the classical Perceptron algorithm and logistic regression using the FOS-GRN model. In brief, our primary objective is \emph{dynamical}: to learn a threshold Boolean network parameterized by $(W,\theta)$ whose asymptotic behavior contains a prescribed set of fixed points and explicitly avoids additional (spurious) fixed points. We do \emph{not} claim to recover a unique underlying ``ground-truth'' network structure or a unique $(W,\theta)$, since multiple threshold Boolean network parameterizations can share the same set of fixed points. Rather, the goal is to infer one plausible and parsimonious parameterization consistent with the prescribed dynamical constraints. The rest of the paper is organized as follows. Section \ref{sec:DATA-AND-METHODS} presents the proposed loss function and the learning algorithm for threshold Boolean networks with prescribed fixed points, the results are reported in Section \ref{sec:RESULTS}, and the conclusion appears in Section \ref{sec:CONCLUSIONS}.

\section{Methods}
\label{sec:DATA-AND-METHODS}
\subsection{Threshold Boolean networks (TBNs)}
A Threshold Boolean Network (TBN) consists of $n$ nodes $x_1, x_2, \dots, x_n$, each taking binary values in $\{0, 1\}$. The state of each node at time $t+1$ is updated using a threshold rule:

\begin{eqnarray}
x_i(t+1) = \begin{cases}
        1 & \text{if } \sum_{j=1}^{n} W_{ji} x_j(t) \geq \theta_i \\
        0 & \text{otherwise}
    \end{cases}
\end{eqnarray}

where $W \in \mathbb{R}^{n \times n}$ is the weight matrix and $\theta \in \mathbb{R}^n$ is the threshold vector.

The updating rule can be written compactly as:
\begin{eqnarray}
x(t+1) = H(x(t) \cdot W - \theta)
\end{eqnarray}

where $H$ is the Heaviside step function applied element-wise.
\paragraph{Fixed-point constraints on $(W,\theta)$.}
For a state $x \in \{0,1\}^n$ to be a fixed point of a TBN, it must satisfy $H(xW+b)=x$ (with $b=-\theta$). This condition translates into a set of per-node inequality constraints on the parameters: for each node $i$, if $x_i=1$ then $(xW)_i + b_i \ge 0$, and if $x_i=0$ then $(xW)_i + b_i < 0$. Therefore, enforcing a prescribed collection of fixed points restricts the feasible region of $(W,b)$; additionally discouraging spurious fixed points further tightens these constraints. In this work, L1 regularization is used to select sparse (and thus more interpretable) solutions among the feasible parameterizations.

\subsection {FOS-GRN model}
The Boolean FOS-GRN model presented in \cite{SANCHEZCORRALES2010971}, consists of thirteen genes: AG, AP1, AP2, AP3, EMF1, FT, FUL, LFY, PI, SEP, TFL1, UFO, WUS. When the model is updated in parallel several times, all the $2^{13} = 8192$ configurations converge to one of the ten fixed points shown in Table \ref{tab2}. The first row of the Table \ref{tab2} shows the names given to the attractors, which correspond to the expression patterns observed in different regions of the developing flower. INF1 = inflorescence attractor 1, INF2 = inflorescence attractor 2, INF3 = inflorescence attractor 3, INF4 = inflorescence attractor 4, SEP = sepal attractor, PET1 = petal attractor 1, PET2 = petal attractor 2, STM1 = stamen attractor 1, STM2 = stamen attractor 2, CAR = carpel attractor.

\begin{table}[!t] \small
\centering
\caption{\textbf{FOS-GRN fixed points}.
    The first row shows the names given to the attractors.}
\scriptsize
\begin{tabular}{ccccccccccc} \hline
 & INF1 & INF2 & INF3 & INF4 & SEP & PET1 & PET2 & STM1 & STM2 & CAR\\ \hline
AG& 0& 0& 0& 0& 0& 0& 0& 1& 1& 1\\
AP1& 0& 0& 0& 0& 1& 1& 1& 0& 0& 0\\
AP2& 0& 0& 0& 0& 1& 1& 1& 1& 1& 1\\
AP3& 0& 0& 0& 0& 0& 1& 1& 1& 1& 0\\
EMF1& 1& 1& 1& 1& 0& 0& 0& 0& 0& 0\\
FT& 0& 0& 0& 0& 1& 1& 1& 1& 1& 1\\
FUL& 0& 0& 0& 0& 0& 0& 0& 1& 1& 1\\
LFY& 0& 0& 0& 0& 1& 1& 1& 1& 1& 1\\
PI& 0& 0& 0& 0& 0& 1& 1& 1& 1& 1\\
SEP& 0& 0& 0& 0& 1& 1& 1& 1& 1& 1\\
TFL1& 1& 1& 1& 1& 0& 0& 0& 0& 0& 0\\
UFO& 0& 1& 0& 1& 0& 1& 0& 1& 0& 0\\
WUS& 0& 0& 1& 1& 0& 0& 0& 0& 0& 0\\\hline
\end{tabular}
\label{tab2}

\end{table}

\subsection{Dataset and learning targets}
The experiments in this paper do not rely on a time-series gene-expression dataset. Instead, the supervision signal is the prescribed set of fixed points of the reference Boolean model. Concretely, the ``dataset'' is the set $\mathcal{F}$ of $m=10$ binary configurations in Table~\ref{tab2}, each configuration being a 13-dimensional vector over the genes (AG, AP1, \dots, WUS).

For training, each fixed point $x \in \mathcal{F}$ is used as an input state at time $t$, and the desired output is the next state at time $t\!+\!1$. Because these states are fixed points, the target satisfies $x(t\!+\!1)=x(t)=x$. Therefore, the supervised pairs are simply $(x, x)$ for all $x \in \mathcal{F}$, where the input is the 13-gene state at time $t$ and the output is the 13-gene state at time $t\!+\!1$.

For the baseline methods (Perceptron and Logistic Regression), we train one model per node $i$ using the same inputs $x \in \mathcal{F}$ and binary labels given by the corresponding component $x_i$, i.e., $x \mapsto x_i$. After training, all methods are evaluated under hard-threshold dynamics by exhaustively enumerating all $2^{13}$ states and collecting those satisfying $f(x)=x$ (as described in Section~\ref{sec:DATA-AND-METHODS}).

\subsection{Loss function construction for learning TBN with prescribed fixed points}
To learn a TBN that contains a specific set of fixed points $\mathcal{F} = \{x^{(1)}, x^{(2)}, \dots, x^{(m)}\}$ and avoids all others, we must define a loss function such that 1) all given fixed points $x \in \mathcal{F}$ satisfy $f(x) = x$, 2) no other states $x \notin \mathcal{F}$ are fixed points, 3) the outputs $f(x)$ are pushed toward binary values, and 4) the model remains sparse through L1 regularization. Considering the four points, we define the following composite loss function:
\begin{eqnarray}
\mathcal{L}(W, b) = \mathcal{L}_\text{pos} + \lambda_\text{neg} \cdot \mathcal{L}_\text{neg} + \lambda_\text{bin} \cdot \mathcal{L}_\text{bin} + \lambda_{\ell_1} \cdot \mathcal{L}_{\ell_1}
\end{eqnarray}

\begin{itemize}
    \item $\mathcal{L}_\text{pos}$ ensures that each desired fixed point $x \in \mathcal{F}$ satisfies $f(x) \approx x$:
    \begin{eqnarray}
        \mathcal{L}_\text{pos} = \frac{1}{|\mathcal{F}|} \sum_{x \in \mathcal{F}} \|\sigma(xW + b) - x\|^2
   \end{eqnarray}
    \item $\mathcal{L}_\text{neg}$ penalizes other spurious fixed points found dynamically:
    \begin{eqnarray}
        \mathcal{L}_\text{neg} = \frac{1}{|\mathcal{S}|} \sum_{x \in \mathcal{S}} \frac{1}{\|\sigma(xW + b) - x\|^2 + \varepsilon}
    \end{eqnarray}
    where $\mathcal{S}$ is the set of currently observed spurious fixed points.
\paragraph{Dynamic update of the spurious set $\mathcal{S}$.}
At training epoch $k$, we define the current set of spurious fixed points as $\mathcal{S}_k = \mathcal{F}^*_k \setminus \mathcal{F}$, where $\mathcal{F}$ is the prescribed set of fixed points (Table~\ref{tab2}) and $\mathcal{F}^*_k$ is the set of fixed points induced by the current parameters $(W,b)$ under the training-time map. In our implementation for $n=13$, $\mathcal{F}^*_k$ is computed by exhaustively enumerating all $2^n$ binary states $x \in \{0,1\}^n$ and collecting those satisfying $\mathrm{round}(\sigma(xW+b)) = x$. The loss term $\mathcal{L}_{\text{neg}}$ is then evaluated on $\mathcal{S}_k$ and updated at the chosen frequency. If $\mathcal{S}_k=\emptyset$, we set $\mathcal{L}_{\text{neg}}=0$ for that epoch.
\newline

    \item $\mathcal{L}_\text{bin}$ encourages binarized outputs:
   \begin{eqnarray}
        \mathcal{L}_\text{bin} = \frac{1}{mn} \sum_{i=1}^{m} \sum_{j=1}^{n} \hat{y}_{ij}(1 - \hat{y}_{ij})
    \end{eqnarray}
    \item $\mathcal{L}_{\ell_1}$ encourages sparsity of weights:
   \begin{eqnarray}
        \mathcal{L}_{\ell_1} = \sum_{i,j} |W_{ij}|
    \end{eqnarray}
\end{itemize}

Here, $\sigma(z)$ is the sigmoid function used only during training for gradient-based optimization, and $b = -\theta$ is the bias vector.

\subsection{Learning algorithm}

The learning algorithm to train a TBN with prescribed fixed points is as follows,
\begin{enumerate}
    \item Initialize $W$ and $b$ randomly.
    \item For each epoch:
    \begin{enumerate}
        \item Compute the sigmoid-based predictions $\hat{y} = \sigma(xW + b)$ for $x \in \mathcal{F}$.
        \item Identify spurious fixed points by computing $\mathcal{F}^*_k$ and setting $\mathcal{S}_k=\mathcal{F}^*_k\setminus\mathcal{F}$. In the present experiments ($n=13$), this is done by exhaustive enumeration of all $2^{13}=8192$ states and checking $\mathrm{round}(\sigma(xW+b))=x$. This search can be executed every epoch, or every $k$ epochs to reduce overhead.
        \item Compute $\mathcal{L}_\text{pos}$, $\mathcal{L}_\text{neg}$, $\mathcal{L}_\text{bin}$, and $\mathcal{L}_{\ell_1}$.
        \item Update $W$ and $b$ using gradient descent or stochastic gradient descent (e.g. Adam) optimizer.
    \end{enumerate}
    \item Stop when convergence criteria or early stopping is met.
\end{enumerate}

\subsection{Validation using traditional TBN evaluation}
After training is complete, we validate the model using a traditional TBN evaluation procedure, i.e., using hard-threshold (Heaviside) dynamics:
 \begin{eqnarray}
    x(t+1) = H(x(t) \cdot W + b)
\end{eqnarray}

To determine the set of fixed points $\mathcal{F}^*$ found by the network:
\begin{enumerate}
    \item Generate all $2^n$ binary states $x \in \{0,1\}^n$.
    \item For each $x$, compute $f(x) = H(xW + b)$.
    \item Collect all $x$ such that $f(x) = x$.
\end{enumerate}

We then compare $\mathcal{F}^*$ with the desired $\mathcal{F}$ to check for completeness and absence of spurious fixed points. Some important points to consider are that during training, sigmoid activations allow differentiability, but at inference time, the network uses only binary updates. Also, the model generalizes well to small and medium networks (e.g., $n \leq 13$) due to exhaustive validation. For larger networks it is recommended to avoid evaluating spurious fixed points on every epoch, instead, only do it every 10 or 20 epochs, or only in the final verification.

\subsection{Scalability and limitations}
A key computational bottleneck of the proposed approach is the identification and verification of spurious fixed points. In our experiments ($n=13$), we can exhaustively enumerate all $2^{n}$ states to (i) detect spurious fixed points during training and (ii) validate the final model under hard-threshold dynamics. However, exhaustive enumeration scales exponentially with $n$ and becomes infeasible for larger networks. For validation when exhaustive enumeration is infeasible, one can approximate fixed-point discovery by sampling large batches of random initial states and simulating the hard-threshold dynamics until convergence, then collecting unique terminal states and verifying the fixed-point condition $f(x)=x$ on those candidates. This trajectory-based validation may miss rare attractors with small basins, but it provides a practical approximation for larger $n$ when full enumeration is not possible.

For larger networks, the training procedure can be adapted by replacing exhaustive spurious-point detection with approximate strategies, such as (a) checking for spurious fixed points only every $k$ epochs, (b) sampling a subset of states and/or sampling trajectories from random initial conditions to discover candidate spurious attractors, and (c) using local search or constraint-based routines to find counterexamples to the fixed-point condition $f(x)=x$. These approaches would allow $\mathcal{L}_{\text{neg}}$ to be computed from a (possibly evolving) subset of spurious candidates, trading completeness for scalability. In addition, the evaluation of candidate states is parallel and can benefit from straightforward parallelization.

Another limitation is that the current loss function explicitly targets fixed points, but it does not directly penalize spurious \emph{limit cycles}. While this is appropriate when fixed points are the main modeling objective, extensions that incorporate multi-step consistency or explicit cycle-avoidance penalties could provide stronger control of the full attractor landscape. Finally, because multiple parameterizations $(W,\theta)$ may share the same fixed points, the method aims to infer a plausible and parsimonious solution (encouraged by L1 regularization), rather than guaranteeing recovery of a unique ``true'' network.

\subsection{Experiment setup}

To evaluate the effectiveness of the proposed learning algorithm for TBNs, we conducted a comparative study against three baseline methods: the standard Perceptron learning algorithm, and Logistic Regression with L1 and L2 regularization, respectively. Each method was run independently over 30 trials. For every run, the model was trained using the same set of 10 known fixed points from Table~\ref{tab2} as supervision. In the proposed method, a custom differentiable loss function was used that incorporates fixed point preservation, dynamic spurious attractor penalization, output binarization, and L1 regularization on weights. In contrast, the Perceptron and Logistic Regression were trained independently for each node using a one-vs-rest scheme. For the baseline methods, the training set consists of the $m=10$ prescribed fixed points $x \in \mathcal{F}$ (Table~\ref{tab2}). For each gene/node $i$, a separate binary classifier is trained using as input the full 13-dimensional state $x$ at time $t$ and as target the $i$-th component $x_i$ (equivalently, the desired next-state component at time $t\!+\!1$ since $x(t\!+\!1)=x(t)$ for fixed points). The learned weight vector for node $i$ defines the $i$-th column of $W$, and the learned intercept defines $b_i$. For Logistic Regression, the parameters are obtained by minimizing the standard cross-entropy objective with either L2 or L1 regularization; for the Perceptron, parameters are obtained via the classical Perceptron update rule. After training, the resulting $(W,b)$ are evaluated under hard-threshold dynamics by enumerating all $2^{13}$ states and collecting those satisfying $f(x)=x$. After training, all models were evaluated by computing the full set of fixed points under hard-threshold (Heaviside) dynamics. We report the total number of fixed points found, the number of correctly matched (desired) fixed points, and the number of spurious fixed points. This experimental setup allows for a direct comparison of how well each method can enforce global dynamical constraints in a data-driven TBN setting.

The main set of experiments employed the following training parameters, determined empirically through preliminary exploration: the number of epochs was set to 1000, with a learning rate of 0.1. The weighting coefficients in the custom loss were fixed as $\lambda_\text{bin} = 0.1$, $\lambda_{\ell_1} = 0.01$, and $\lambda_\text{neg} = 5.0$. The numerical stability constant was set to $\varepsilon = 10^{-6}$, and the network size was $n = 13$, corresponding to the number of genes considered. These values provided stable convergence and satisfactory separation between desired and spurious attractors. In our implementation for the reported experiments ($n=13$), the dynamic search for spurious fixed points (and thus the update of $\mathcal{S}_k$) was executed at every epoch; this step dominates the computational cost of the proposed method, but remains tractable at this network size.

A second experiment was conducted to investigate the effect of the L1 regularization coefficient ($\lambda_{\ell_1}$) on the learned network topology. Specifically, the proposed method was retrained 30 times for each of the following $\lambda_{\ell_1}$ values: 0, 0.001, 0.005, 0.01, 0.05, 0.1, 0.5, 1, 5, and 10. For each configuration, we monitored the number of successful networks (i.e., those achieving exactly the 10 desired fixed points and no spurious ones) and analyzed the corresponding network structures.

The outcomes of both experiments, including comparative performance with baseline models and the robustness analysis of the proposed approach, are presented and discussed in Section~\ref{sec:RESULTS}.

It is important to point out that the goal of this study is to provide a first proof-of-concept of the proposed loss-driven learning framework for TBNs with prescribed fixed points. We therefore focus on the widely used FOS-GRN model of \textit{Arabidopsis thaliana} as a well-established reference system with a known set of biologically meaningful fixed points and documented challenges for threshold-based inference. In addition, we include a qualitative consistency check by comparing dominant signed interactions in the inferred TBN (Fig.~\ref{fig:TBNexample1}) with the regulatory logic reported for the original FOS-GRN model in \cite{SANCHEZCORRALES2010971}. While additional biological networks and synthetic benchmarks would further support generalizability claims, such an extended benchmark study is left for future work.

\section{Results}
\label{sec:RESULTS}

The results of the 30 independent runs are summarized in Table \ref{tab:tbn_comparison_detailed}. We notice that the proposed method with the custom loss function was the only method capable of inferring networks (five) with the desired fixed points and no spurious fixed points. On average it found 8 of the 10 fixed points, and in none of the runs the learned networks contained spurious fixed points. On average, the Perceptron produced networks with 26 fixed points, of which 10 matched the desired ones and 16 were spurious. This is consistent with what was observed in \cite{10944722}. When using Logistic Regression (L2), no dispersion was observed within the 30 runs, it always obtained the same result, that is, networks with 18 fixed points, 10 of them matching the desired fixed points and the remaining 8 were spurious fixed points. Finally, the worst results were obtained for Logistic Regression (L1), with high dispersion in the results, with on average, inferring networks with 38 fixed points, where 7 matched the desired fixed points and the remaining 31 were spurious fixed points. The proposed method had the largest run time, with an average of 19 seconds per run. As explained previously, this could be optimized by not evaluating spurious fixed points on every epoch, instead, only do it every 10 or 20 epochs.

\begin{table}[t]
\centering
\caption{\textbf{Experimental results}. Comparison of TBN learning methods over 30 independent runs. A run is considered successful when it recovers \emph{exactly} the 10 desired fixed points and no spurious fixed points.}
\label{tab:tbn_comparison_detailed}
\tiny
\setlength{\tabcolsep}{2.5pt}
\renewcommand{\arraystretch}{1.15}

\begin{tabular}{|l|c|c|c|c|c|}
\hline
\textbf{Method} &
\textbf{Success}/\textbf{Runs} &
\shortstack{\textbf{Fixed Points}\\($\bar{x}\pm$ sd)} &
\shortstack{\textbf{Matched}\\($\bar{x}\pm$ sd)} &
\shortstack{\textbf{Spurious}\\($\bar{x}\pm$ sd)} &
\shortstack{\textbf{Time}\\(s)} \\
\hline
Custom Loss & 5/30 & 8.53 $\pm$ 0.90 & 8.53 $\pm$ 0.90 & 0.00 $\pm$ 0.00 & 19.68 $\pm$ 0.84 \\
Perceptron & 0/30 & 26.50 $\pm$ 5.72 & 9.83 $\pm$ 0.38 & 16.67 $\pm$ 5.65 & 0.0045 $\pm$ 0.0026 \\
Logistic Regression (L2) & 0/30 & 18.00 $\pm$ 0.00 & 10.00 $\pm$ 0.00 & 8.00 $\pm$ 0.00 & 0.0118 $\pm$ 0.0069 \\
Logistic Regression (L1) & 0/30 & 38.47 $\pm$ 20.39 & 7.47 $\pm$ 1.04 & 31.00 $\pm$ 20.37 & 0.0047 $\pm$ 0.0042 \\
\hline
\end{tabular}
\end{table}

\begin{figure}
\centering
\includegraphics[scale=0.21]{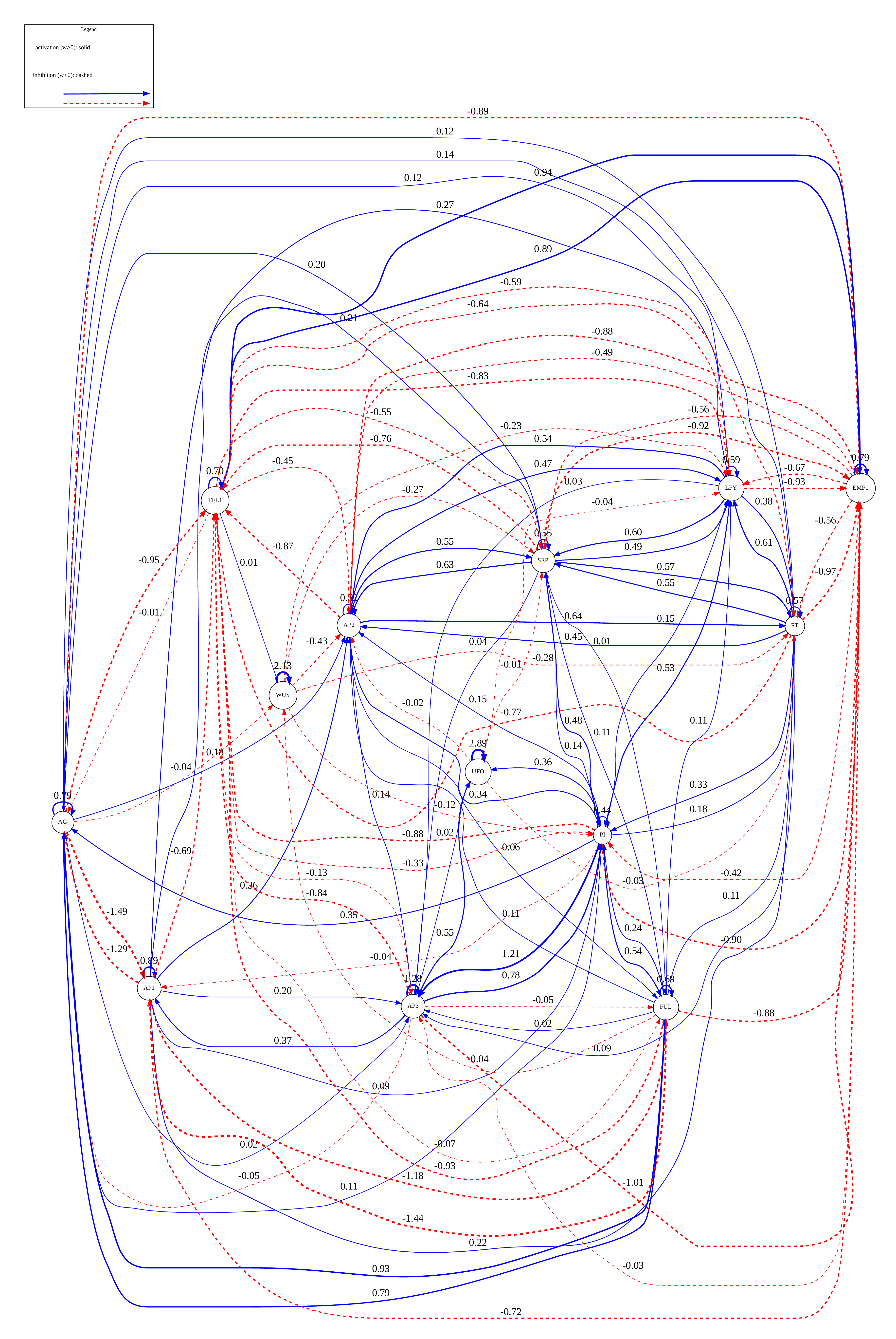}
\caption{\textbf{A TBN successfully inferred by the proposed approach.} Edge labels indicate weights; solid edges denote activation ($w>0$) and dashed edges denote inhibition ($w<0$); edge thickness is proportional to $|w|$.}
\label{fig:TBNexample1}
\end{figure}

\begin{figure}
\centering
\includegraphics[scale=1.1,trim=4cm 0cm 0cm 0cm,clip]{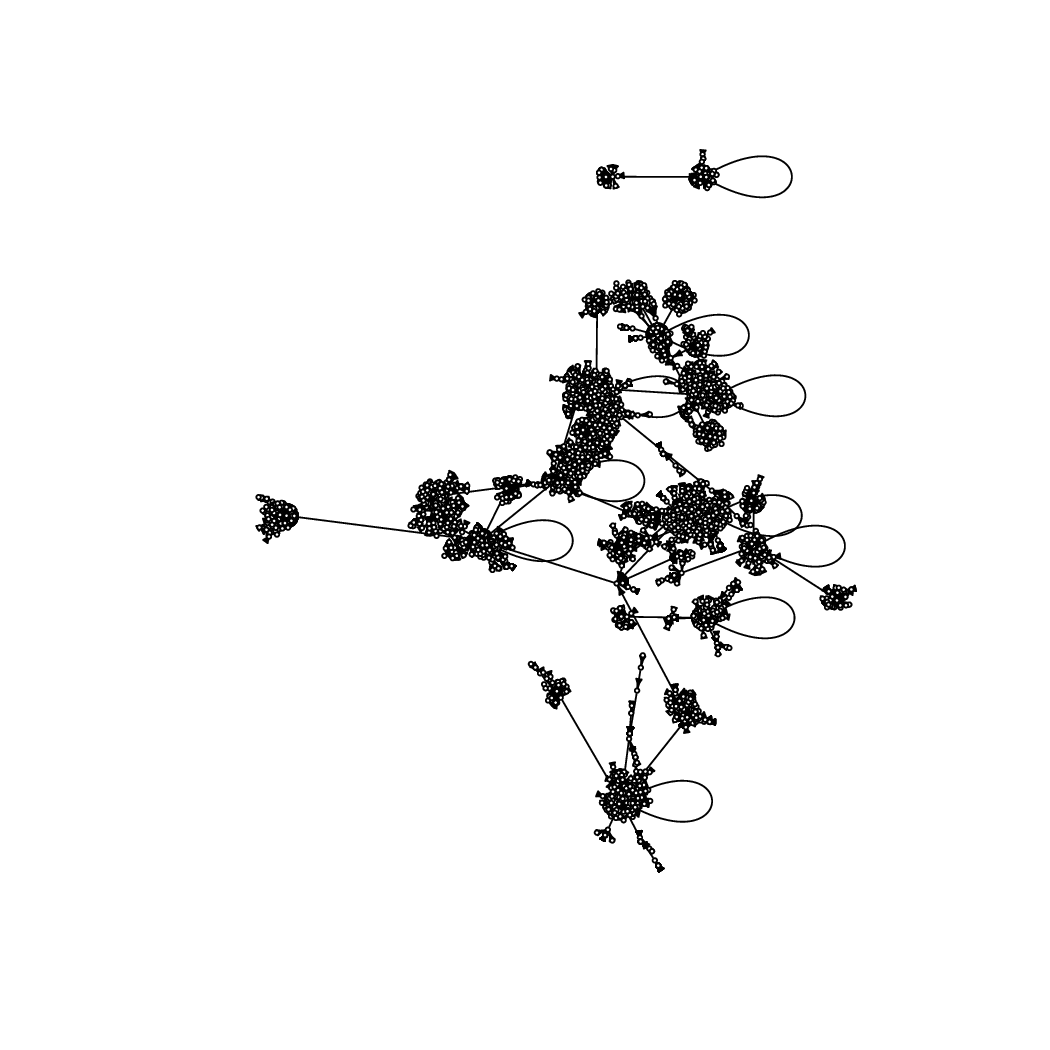}
\caption{\textbf{The state transition graph of the TBN of Fig.~\ref{fig:TBNexample1}.} The fixed points are represented by nodes with a self-loop. }
\label{fig:STG1}
\end{figure}

An example of one of the five successful TBN inferred by the proposed approach is shown in Fig. \ref{fig:TBNexample1}, and its respective state transition graph in Fig. \ref{fig:STG1}, where the fixed points are identified by a self-loop. Notice that there are exactly 10 fixed points. 

\begin{figure}[t]
\centering
\includegraphics[width=1\textwidth]{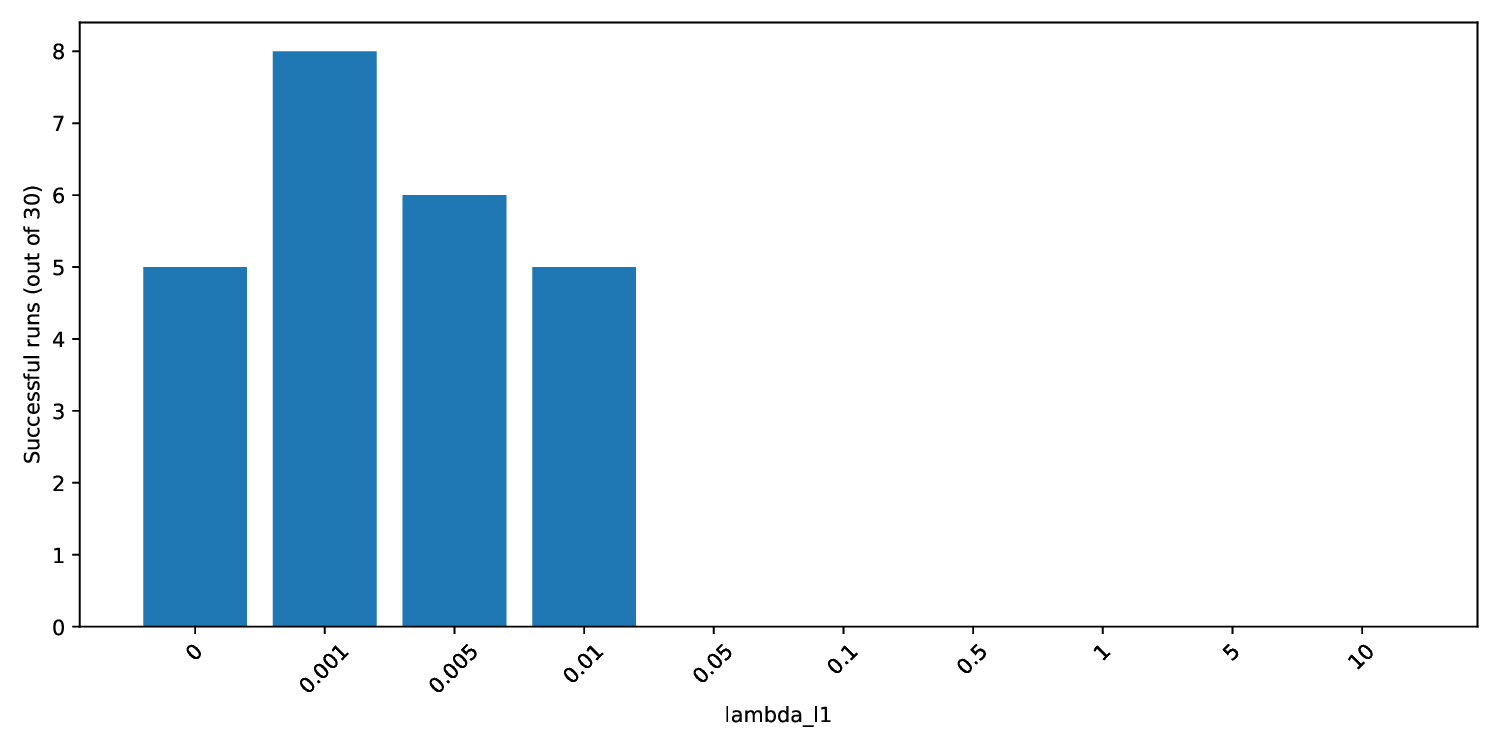}
\caption{\textbf{Effect of L1 regularization on model performance.} Number of successful networks (out of 30 runs) as a function of the L1 regularization coefficient ($\lambda_{\ell_1}$).}
\label{fig:lambda_sweep}
\end{figure}

In addition to these experiments, we further investigated the effect of varying the L1 regularization coefficient ($\lambda_{\ell_1}$) on the learning performance of the proposed method. For this purpose, the model was retrained 30 times for each $\lambda_{\ell_1}$ value in $\{0, 0.001, 0.005, 0.01, 0.05, 0.1, 0.5, 1, 5, 10\}$. Figure~\ref{fig:lambda_sweep} summarizes the number of successful networks (i.e., those achieving the 10 desired fixed points with no spurious ones) obtained for each value of $\lambda_{\ell_1}$. The results show that the number of successful runs remained approximately constant across the regularization strengths, from $\lambda_{\ell_1}$ value in $\{0, 0.001, 0.005, 0.01\}$, with $\lambda_{\ell_1}= 0.001$ obtaining the best results (8 in total). No systematic effect was observed on the average in-degree of the successful networks. Then for $\lambda_{\ell_1}\geq 0.05$ no successful networks were found. 

\subsection{Qualitative comparison with the logical FOS-GRN model}
Here we report this comparison for a representative successful inferred network (Fig.~\ref{fig:TBNexample1}); a systematic consensus analysis of interaction frequencies across multiple successful runs is left for future work.

Although the main objective of this work is methodological (enforcing a prescribed set of fixed points), we provide a qualitative consistency check between the inferred TBN in Fig.~\ref{fig:TBNexample1} and the logical FOS-GRN model reported in \cite{SANCHEZCORRALES2010971}. In the inferred TBN, the sign of $W_{ji}$ indicates whether node $j$ tends to activate ($W_{ji}>0$) or inhibit ($W_{ji}<0$) node $i$, while larger $|W_{ji}|$ suggests stronger relative influence within the threshold update mechanism.

Several dominant signed interactions in Fig.~\ref{fig:TBNexample1} are consistent with key monotone relationships encoded in the logical model. For instance, the inferred network captures the antagonism between AG and AP1 through strong negative cross-regulation ($W_{\text{AP1}\to\text{AG}}=-1.29$ and $W_{\text{AG}\to\text{AP1}}=-1.49$). It also reflects the inhibitory role of EMF1 on flowering activators via negative connections to FT and LFY ($W_{\text{EMF1}\to\text{FT}}=-0.56$ and $W_{\text{EMF1}\to\text{LFY}}=-0.67$), and the inhibitory role of TFL1 through multiple negative outgoing edges (e.g., to LFY, FT, AP2 and SEP). Moreover, the inferred PI module exhibits strong positive regulation by AP3, LFY, SEP and self-maintenance, in agreement with the logical rules involving these components.

Regarding UFO, the logical model treats it as an unregulated/persistent component (its value is preserved across updates for a given initial condition). Consistently, the inferred TBN assigns UFO a dominant positive self-loop ($W_{\text{UFO}\to\text{UFO}}=2.89$), which promotes state preservation; additional incoming terms may arise because the present training objective targets fixed points rather than exact reconstruction of the full transition table.

Overall, this comparison suggests that the proposed learning approach can recover a sparse TBN whose dominant signed interactions are broadly aligned with the regulatory logic of the reference model, while still emphasizing that multiple parameterizations $(W,\theta)$ may realize the same fixed points and that threshold dynamics can only approximate highly context-dependent logical interactions.

\section{Conclusion}
\label{sec:CONCLUSIONS}
We introduced a learning algorithm for threshold Boolean networks that successfully enforces a prescribed set of fixed points while avoiding spurious ones. Our approach offers superior control over network asymptotic behavior compared to classical methods, with a subset of runs achieving perfect reconstruction of all desired attractors. This makes the method particularly suitable for modeling gene regulatory systems with known stable states. Regarding scalability, the current implementation relies on exhaustive state enumeration to detect and verify spurious fixed points, which is feasible for small networks but grows exponentially with $n$. Future work will investigate approximate and constraint-based alternatives for spurious attractor discovery, as well as extensions to explicitly control spurious limit cycles in larger-scale settings. Also we plan to apply this framework to other biological networks of varying complexity, including models of metabolic and signaling pathways, to further assess its generalizability and scalability.

\section*{Code availability}
To support reproducibility, the implementation of the proposed learning algorithm and the scripts used to run the experiments are publicly available at: \texttt{https://github.com/gruzh/LA4TBN\_PFP}.

\begin{credits}
\subsubsection{\ackname} The author thanks ANID FONDECYT 1230315, ANID-MILENIO-NCN2024\_103, ANID-MILENIO-NCN2024\_047, ANID PIA/BASAL AFB240003, and Centro de Modelamiento Matemático (CMM) FB210005, BASAL funds for centers of excellence from ANID-Chile.

\subsubsection{\discintname}
The author has no competing interests to declare that are relevant to the content of this article.
\end{credits}

%
%
%
\bibliographystyle{splncs04}
\bibliography{bibliography_CIBB_file}

\end{document}